\documentclass[letterpaper]{article} 
\usepackage[preprint]{aaai2027}  
\usepackage[hyphens]{url}  
\usepackage{graphicx} 
\usepackage{natbib}  
\usepackage{caption} 
\usepackage{algorithm}
\usepackage{algorithmic}
\usepackage{amsmath,amssymb}
\usepackage{array}
\usepackage{newfloat}
\usepackage{listings}
\DeclareCaptionStyle{ruled}{labelfont=normalfont,labelsep=colon,strut=off} 
\floatstyle{ruled}
\newfloat{listing}{tb}{lst}{}
\floatname{listing}{Listing}

\usepackage{booktabs}

\title{Cross-Visibility-Gap Identity Persistence for 4D Scene Graphs}
\author{
   Jiahui Hu\textsuperscript{\rm 1},
   Li Gao\textsuperscript{\rm 2},
   Yu Liu\textsuperscript{\rm 3},
   Pengzhan Zhou\textsuperscript{\rm 2}
}
\affiliations{
   \textsuperscript{\rm 1}Department of Computer Science, The University of Hong Kong\\
   \textsuperscript{\rm 2}Chongqing University\\
   \textsuperscript{\rm 3}Department of Computing, The Hong Kong Polytechnic University
}

\begin{document}

\maketitle

\begin{abstract}
4D scene graphs require persistent object identity, but a track may terminate when an object is transported inside an opaque carrier and later reappear as a fresh detection. 
In this regime, target appearance, self-motion, and last-seen geometry cannot determine its historical identity. Our key insight is that identity evidence need not remain attached to the invisible target: it can be transferred to the observable carrier and recovered after the gap. 
We therefore formulate closed-set cross-visibility-gap identity persistence as historical-node recovery through a remembered target--carrier binding. 
This turns a visibility discontinuity into a graph-routing problem: when direct target evidence disappears, identity can persist through a relation anchored to an observable carrier. 
Unlike extrapolative tracking, the method does not hallucinate target motion inside the carrier; it stores the last observed binding and delays identity assignment until reappearance supplies new routing evidence.

We introduce CVG-SGM, a perception-frontend-compatible transfer--maintain--read memory. 
Geometry-weighted cross-node interaction writes target-specific evidence into carrier states, a shared per-node GRU maintains the binding during transport, and a two-time readout combines stored carrier--identity compatibility with current detection--carrier geometry to re-link a fresh detection. 
This separates identity evidence---who was carried---from routing evidence---where the returning detection belongs---and treats graph relations as memory paths rather than only prediction outputs.

In the controlled one-identity-per-carrier regime, CVG-SGM reaches $1.000$ re-link accuracy versus $0.570$ for no-memory and parameter-count-matched no-write controls, and generalizes to scene-disjoint unseen longer gaps. 
On real RGB-D/SAM2 feature-level episodes, it re-links 36/36 held-out detections, compared with 28/36 for the strongest fixed carrier--identity control.
Counterfactual carrier-memory swaps redirect re-linking according to the swapped binding, confirming that stored relational content drives recovery.
\end{abstract}

\section{Introduction}
A temporal scene graph depends critically on persistent object identity: when a node no longer refers to the same physical object, its accumulated relations, attributes, and task history can no longer be reliably associated with that object.
This problem becomes especially acute when an object is placed inside a moving carrier, disappears from view, and later reappears as a new detection. 
During transport, the target's appearance is unavailable and its trajectory is no longer directly observable; if the returning detection is instantiated as a fresh node, it cannot inherit the state accumulated under the original identity. 
Scene graphs and their 3D/4D extensions model objects, relations, geometry, and temporal change for visual reasoning and embodied planning~\cite{johnson2015image,armeni2019scene,yang20234d,rana2023sayplan,gu2024conceptgraphs}.
However, existing 4D scene-graph pipelines primarily maintain identity through continuous tracking, appearance, or short-term geometric association, and do not provide an explicit mechanism for restoring a historical graph identity once a returning detection's observation track has broken after a prolonged visibility gap. 
We refer to this under-addressed capability as \emph{cross-visibility-gap identity persistence}.

\begin{figure*}[!t]
\centering
\includegraphics[width=\textwidth]{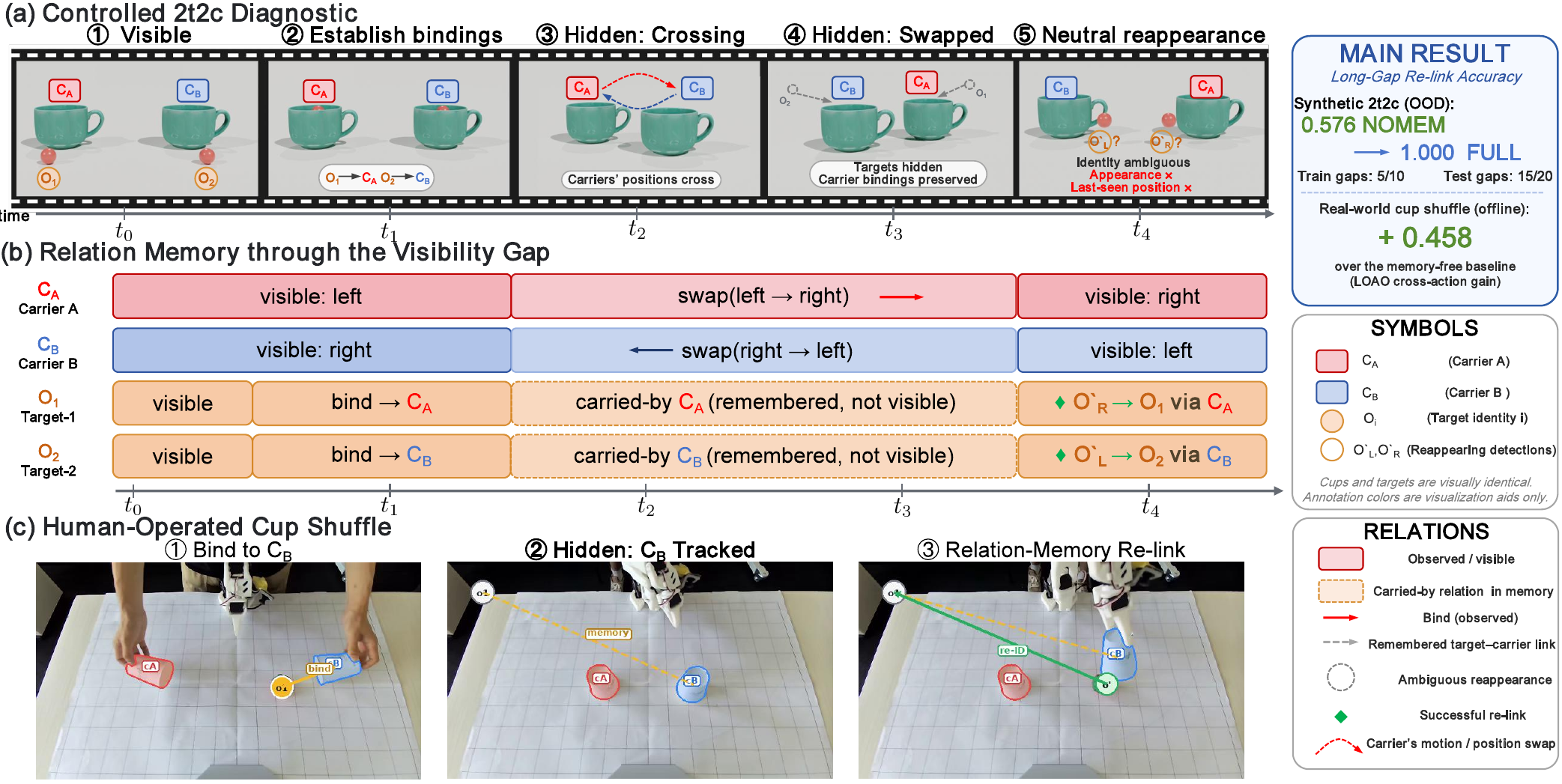}
\caption{
\textbf{Cross-visibility-gap identity persistence through remembered carrier relations.}
\textbf{(a)} Controlled 2t2c diagnostic: two identical targets bind to carriers $c_A$ and $c_B$, are occluded while the carriers swap, and reappear near their current carriers at locations that are uninformative with respect to their own pre-gap positions.
\textbf{(b)} The bindings persist as memory through the gap, re-linking each fresh detection to its identity; on unseen long gaps, FULL improves re-link accuracy from $0.576$ (NOMEM) to $1.000$.
\textbf{(c)} Complementary human-operated cup-shuffle videos provide
offline carrier-recovery evidence, while full real RGB-D/SAM2 feature-level
2t2c re-linking is reported in Table~\ref{tab:real_2t2c}.
}
\label{fig:teaser}
\end{figure*}

A particularly diagnostic case involves indistinguishable objects carried inside identical moving containers. Appearance is uninformative; the target's trajectory cannot be reliably extrapolated from its own history because its motion is governed by the carrier; and even the containment relation may become temporarily unobservable. 
The carrier, however, remains trackable in this diagnostic setting. If the graph records before disappearance that target $o_i$ entered carrier $c_j$, tracks $c_j$ through the gap, and retrieves this binding when the target reappears as a fresh detection, it can re-link that detection to the original target identity---persistence through the remembered object--carrier relation, \emph{who was carried by what}, rather than through target appearance or trajectory.
Our focus is the distinct broken-association setting in which the target track has terminated and a returning detection is instantiated as a fresh node that must be actively re-linked to an existing graph identity.
We introduce \textbf{CVG-SGM}, a perception-frontend-compatible identity
re-linking module built around a transfer--maintain--read cycle. Cross-node
interaction transfers target-specific evidence into observable carrier states,
per-node recurrence maintains it through the visibility gap, and a two-time
readout combines stored identity compatibility with current routing geometry.
Figure~\ref{fig:teaser} illustrates the diagnostic setting and the role of
remembered carrier relations.

Experimentally, FULL reaches $1.000\pm0.000$ in controlled 2t2c, while NOMEM and the parameter-count-matched NOWRITE control remain at $0.570\pm0.000$. A single model trained on 5- and 10-frame gaps retains perfect accuracy on scene-disjoint 15- and 20-frame gaps. 
On real RGB-D/SAM2 feature-level 2t2c episodes, Neural FULL correctly re-links 36/36 held-out detections, compared with 28/36 for the strongest fixed carrier--identity control.
Identity-key permutation, carrier-memory swapping, propagated read, pool-order reversal, and structured corruptions support the same carrier-conditioned binding mechanism. Our instance-level claim is limited to the one-identity-per-carrier regime.

\paragraph{Contributions.}
\begin{itemize}
\item We formulate \emph{cross-visibility-gap identity persistence} as
historical-node recovery when target tracking, appearance, and target-centric
motion are unavailable.

\item We propose \textbf{CVG-SGM}, a transfer--maintain--read memory that stores
target evidence in carrier states and re-links fresh detections through a
two-time memory-and-geometry readout.

\item We isolate the mechanism through parameter-count-matched no-write, identity-permutation, carrier-memory-swap, propagated-read, and pool-order controls, and evaluate it on independently generated canonical scenes, scene-disjoint unseen gaps, structured node-level corruptions, real RGB-D/SAM2 full re-linking, roster scaling, and shared-carrier ambiguity.
\end{itemize}

\section{Related Work}

\subsection{Occlusion Tracking and Object Memory}
Object-permanence and occlusion-tracking methods maintain target continuity during disappearance using appearance cues, propagated masks, or hidden-motion estimates~\cite{shamsian2020learning,tokmakov2021learning,tokmakov2022object,oh2019video,cheng2022xmem}. 
Promptable video trackers and tracking-through-occlusion benchmarks further improve target-centric continuity under prolonged occlusion~\cite{ravi2025sam2,vanhoorick2023tracking}. 
Recent shell-game diagnostics evaluate whether models can preserve the identities of visually indistinguishable entities through spatiotemporal continuity~\cite{liu2026shellgame}, while object-memory systems maintain explicit entity representations for interaction, hidden-object reasoning, or scene-state anticipation~\cite{huang2024outofsight,teng2021target,peddi2024anticipation}. 
In contrast, CVG-SGM addresses a broken-association regime in which the original target track has terminated and a fresh detection must be re-linked to its historical graph node through a remembered target--carrier binding.

\subsection{Temporal, World, and 4D Scene Graphs}
Scene graphs and their 3D/4D extensions model objects, relations, geometry,
and temporal change for visual reasoning and embodied use
~\cite{krishna2017visual,rosinol20203d,yang20234d,rana2023sayplan,gu2024conceptgraphs}.
Dynamic graph and incremental mapping systems maintain identity through
geometric fusion, frontend associations, interaction tracking, or semantic
matching
~\cite{ji2020action,cong2021spatial,wu2021scenegraphfusion,hughes2022hydra,behrens2025lostfound,almdfaa2025vlkng}.
CVG-SGM targets the complementary case where visually indistinguishable objects
reappear as fresh nodes and must be re-linked through carrier-mediated relation
memory.

\section{Method}

\subsection{Problem Formulation}
We model a dynamic scene as a 4D scene graph whose nodes are objects and whose edges are pairwise relations, both indexed by time.
Each node $n$ carries a frontend-provided feature $\mathbf{x}_n^t\in\mathbb{R}^{D}$ containing geometric, observability, and category-related cues, including its image-plane position $\mathbf{p}_n^t$ and a visibility indicator $v_n^t\in\{0,1\}$.

When an object's track is interrupted by a long visibility gap---for example, because it becomes contained in or transported by an observable carrier and is consequently occluded---it may re-enter the scene as a \emph{fresh} detection with no persistent identifier.
The cross-visibility-gap re-linking task is to assign each fresh detection to the historical node it continues.

In our appearance-degenerate diagnostic, visual appearance is uninformative and target-centric motion is unavailable through the gap; recovery must therefore rely on relational history accumulated while the target was associated with an observable carrier.

Let $\mathcal{H}$ denote the set of historical target nodes, $\mathcal{P}$ the set of fresh detections, and $\mathcal{C}$ the set of candidate carriers. 
For each fresh detection $k\in\mathcal{P}$, the model predicts
\begin{equation}
\hat{\imath}_k = \arg\max_{i\in\mathcal{H}} s_{ki},
\label{eq:objective}
\end{equation}
where $s_{ki}$ is the re-link compatibility between fresh detection $k$ and historical identity $i$, defined in the readout below.

\subsection{CVG-SGM Overview}

\begin{figure*}[t]
\centering
\includegraphics[width=\textwidth]{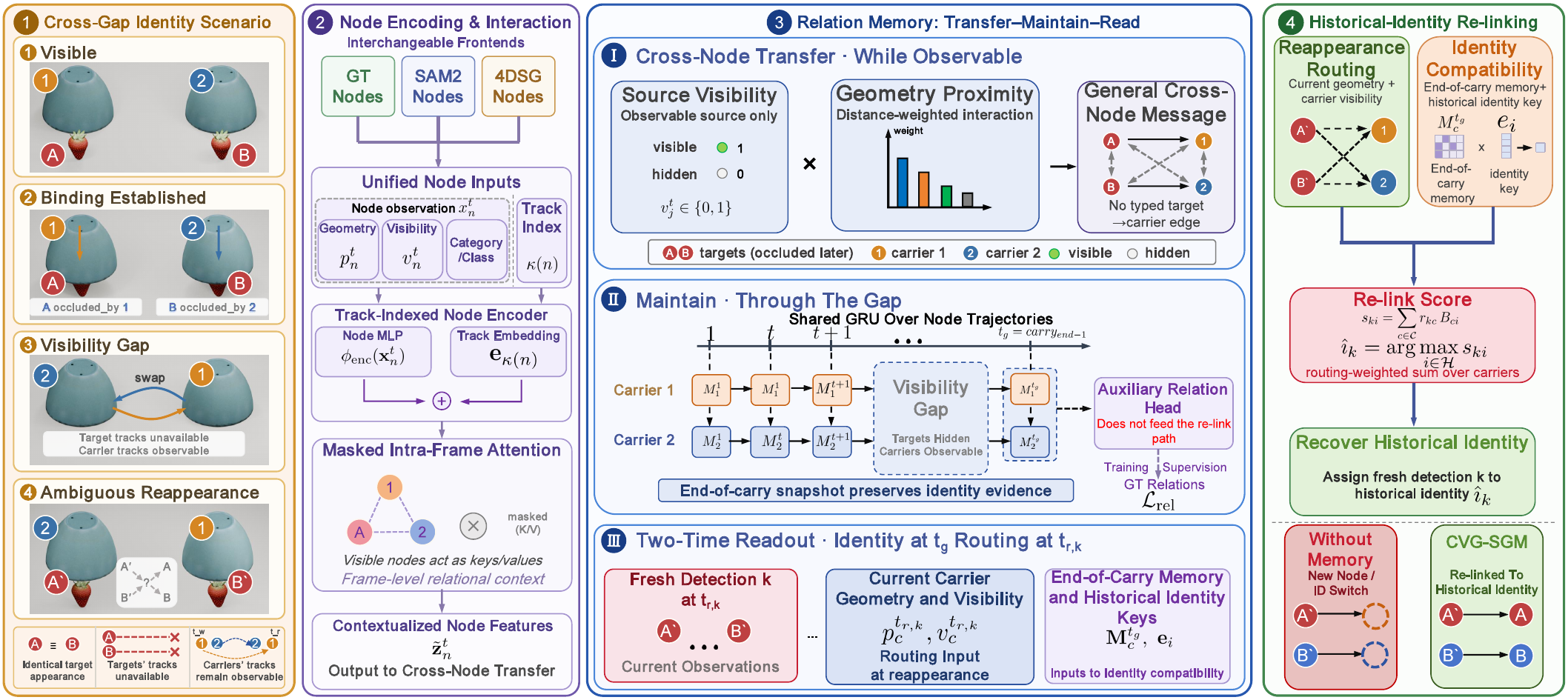}
\caption{
\textbf{Overview of CVG-SGM.}
\textbf{(1)} Identical targets disappear with separately trackable carriers and reappear without continuous target tracks.
\textbf{(2)} Nodes from GT, SAM2, or 4DSG pass through a track-indexed encoder and masked intra-frame attention.
\textbf{(3)} Geometry-weighted cross-node transfer lets target-specific evidence enter a carrier's recurrent state, updated by a shared GRU that maintains it through the gap; an auxiliary relation head reads these states under relation supervision.
\textbf{(4)} At reappearance, routing weights $r_{kc}$ from current detection--carrier geometry combine with identity compatibility $B_{ci}$ from end-of-carry memory as $s_{ki} = \sum_{c\in\mathcal{C}} r_{kc} B_{ci}$ to recover the historical identity.
}
\label{fig:method}
\end{figure*}

As shown in Fig.~\ref{fig:method}, CVG-SGM follows a transfer--maintain--read cycle: geometry-weighted cross-node transfer writes target-specific evidence into an observable carrier state, per-node recurrent states updated by a shared GRU maintain it through the gap, and a two-time readout combines stored carrier--identity compatibility with reappearance geometry. 
Masked intra-frame attention adds an implicit cross-node path; the auxiliary relation logits do not enter the inference-time re-linking, though the relation loss supervises shared representations during training.

\subsection{Node Encoding and Masked Intra-Frame Interaction}

Each node feature is encoded and augmented with a learned track-indexed embedding,
\begin{equation}
    \mathbf{z}_n^t = \phi_{\mathrm{enc}}(\mathbf{x}_n^t)
    +\mathbf{e}_{\kappa(n)}
    \label{eq:node_encoding}
\end{equation}
where $\phi_{\mathrm{enc}}$ is an MLP, $\kappa(n)$ denotes the
episode-local track address assigned to node $n$, and $\mathbf{e}_{\kappa(n)}$ is the corresponding learned embedding. 
These embeddings serve as episode-local graph addresses rather than semantic identity descriptors. 
Historical target-track embeddings are reused as candidate identity keys, while fresh detections receive a shared reserved pool address. 
The intended computation is therefore to transfer address-coded evidence into carrier states and later recover the corresponding historical node; we do not claim appearance-based or open-vocabulary re-identification. 
Re-linking consequently depends on stored carrier memory and current routing geometry rather than a persistent identifier attached to the fresh detection. We evaluate an episode-wise identity-key permutation control to test whether the model exploits a globally fixed identity-slot ordering. 
Visibility $v_n^t$ masks intra-frame interaction.

Within each frame, nodes exchange information through masked multi-head
self-attention. Let $\mathbf{Z}_{\mathrm{vis}}^t = \{\mathbf{z}_j^t \mid v_j^t=1\}$ denote the set of currently visible node features. 
Then,
\begin{equation}
    \tilde{\mathbf{z}}_n^t
    =
    \operatorname{MHA}\left(
        \mathbf{z}_n^t,\,
        \mathbf{Z}_{\mathrm{vis}}^t
    \right).
    \label{eq:masked_attention}
\end{equation}
Currently invisible nodes are excluded from
$\mathbf{Z}_{\mathrm{vis}}^t$ and therefore do not serve as attention keys or
values. This creates an implicit cross-node interaction path complementary to
the explicit geometry-weighted message.

\subsection{Geometry-Weighted Cross-Node Relational Transfer}

In addition to masked attention, CVG-SGM applies a general proximity-weighted message over ordered node pairs, 
\begin{align}
    a_{nj}^t &= \mathbf{1}[n\neq j]\; v_j^t\,\exp\!\Big(-\tfrac{\lVert \mathbf{p}_n^t-\mathbf{p}_j^t\rVert_2}{\tau_w}\Big)
    \label{eq:write_weight}\\
    \mathbf{m}_n^t &= \sum_{j} a_{nj}^t\, W_{\mathrm{write}}\,\tilde{\mathbf{z}}_j^t
    \label{eq:cross_node_message}\\
    \mathbf{g}_n^t &= \tilde{\mathbf{z}}_n^t + \mathbf{m}_n^t 
    \label{eq:memory_input}
\end{align}

The exchange is defined over all ordered node pairs, excludes self-messages, and is gated by the visibility of the \emph{source} $j$, so an invisible node contributes no new message.
The architecture imposes no typed target$\to$carrier edge and no carrier-only receiver mask.

When a target and a carrier are spatially associated before disappearance, target-specific information can enter the carrier state through this generic transfer. 
The resulting carrier-specific content is therefore a task-induced consequence of the learned interaction rather than a typed target-to-carrier architectural constraint. 
This explicit geometry-weighted message provides a second cross-node transfer path in addition to masked intra-frame attention.

\subsection{Recurrent Memory and Auxiliary Relation Prediction}

A shared GRU is applied independently along each node trajectory to integrate the contextualized features and geometry-weighted messages over time,
\begin{equation}
    \mathbf{M}_n^{1:T} = \mathrm{GRU}\!\big(\mathbf{g}_n^{1:T}\big),
    \label{eq:recurrent_memory}
\end{equation}

This yields one recurrent state $\mathbf{M}_n^t$ per node while sharing the GRU parameters across nodes. The recurrent state maintains transferred information after the target's observation track becomes unavailable.

In the NOMEM memory-free ablation, both the explicit geometry-weighted message and the recurrent GRU update are bypassed. 
The contextualized per-frame representation $\tilde{\mathbf{z}}_n^t$ is used directly in place of the recurrent state, while masked intra-frame attention is retained.

An auxiliary relation head reads the recurrent node states,
\begin{equation}
    \boldsymbol{\ell}^{t}
    =
    \phi_{\mathrm{rel}}(\mathbf{M}^{t})
    \in
    \mathbb{R}^{N\times N\times|\mathcal{R}|},
    \label{eq:relation_head}
\end{equation}
where $N$ is the number of nodes and $\boldsymbol{\ell}_{nj}^{t}\in\mathbb{R}^{|\mathcal{R}|}$ contains the logits for ordered node pair $(n,j)$ over relation categories $\mathcal{R}$. These logits are supervised by binary cross-entropy against the ground-truth pairwise relation labels.

The predicted relation logits do not gate the cross-node transfer and do not enter the identity re-linking score at inference time. 
The auxiliary relation loss nevertheless backpropagates through the recurrent node states and shared upstream representations, providing additional supervision during training.

\subsection{Two-Time Carrier-Memory Readout}

Re-linking draws on two decoupled sources of evidence observed at different
frames.~\textbf{Identity evidence} is read from the canonical end-of-carry
carrier-memory snapshot, denoted by $t_g$, through the bilinear compatibility
\begin{equation}
    B_{ci} = \mathrm{Bil}\!\big(\mathbf{M}_c^{t_g},\,\mathbf{e}_i\big)
       = (\mathbf{M}_c^{t_g})^{\top}\, W_{\mathrm{bil}}\, \mathbf{e}_i
    \label{eq:compat}
\end{equation}
between carrier $c$'s recurrent memory $\mathbf{M}_c^{t_g}$ and the identity key
$\mathbf{e}_i$ of historical target $i$, where $W_{\mathrm{bil}}$ is a learned
weight matrix. A higher $B_{ci}$ indicates that carrier $c$'s memory more
strongly encodes a binding to target $i$---i.e.\ that $c$ is more likely the
carrier that held target $i$ during the gap.
In the controlled diagnostic, $t_g$ is provided by the scripted end-of-carry event and is used only as a canonical analysis snapshot. The inference rule need not know this event boundary: replacing $\mathbf{M}_c^{t_g}$ with the latest pre-reappearance state~$\mathbf{M}_c^{t_{r,k}-1}$ produces identical predictions in the propagated-read diagnostic.

\textbf{Routing evidence} is obtained from the geometry at fresh detection $k$'s reappearance frame $t_{r,k}$,
\begin{align}
    d_{kc} &= \lVert \mathbf{p}_k^{t_{r,k}}-\mathbf{p}_c^{t_{r,k}}\rVert_2 + \lambda_{\mathrm{vis}}\,(1-v_c^{t_{r,k}})
    \label{eq:routing_distance}\\
    r_{kc} &= \frac{\exp(-d_{kc}/\tau_a)}{\sum_{c'}\exp(-d_{kc'}/\tau_a)}
    \label{eq:routing}
\end{align}
where invisible carriers incur the fixed penalty $\lambda_{\mathrm{vis}}=10$.

The two evidence sources are combined through a routing-weighted aggregation,
\begin{equation}
    s_{ki}
    =
    \sum_{c\in\mathcal{C}} r_{kc}\,B_{ci},
    \label{eq:score}
\end{equation}
where the normalized routing weights determine how strongly the identity compatibility stored in each candidate carrier contributes to fresh detection
$k$. 
The fresh detection is then re-linked to the historical node maximizing $s_{ki}$ according to Eq.~\eqref{eq:objective}.

The predicted historical-node index provides the re-linking interface through which a downstream graph manager can attach the fresh observation track to an existing node. We evaluate recovery of this historical identity; executing the subsequent online graph update is outside the scope of this work.

\subsection{Training Objective and Complexity}

The canonical re-linking model is trained with
\begin{equation}
    \mathcal{L} = \mathcal{L}_{\mathrm{id}} + \mathcal{L}_{\mathrm{rel}}
    \label{eq:total_loss}
\end{equation}
where $\mathcal{L}_{\mathrm{id}}$ is cross-entropy over the historical-identity scores $s_{ki}$ against the ground-truth slot, and $\mathcal{L}_{\mathrm{rel}}$ is binary cross-entropy over pairwise relation logits.

For the latency benchmark configuration ($N{=}13$, $T{=}30$, $d_{\mathrm{hid}}{=}128$), the shared GRU adds $99{,}072$ parameters ($+56.6\%$) and raises median forward latency from $1.752$ to $2.150$~ms.
Online inference stores $O(Nd_{\mathrm{hid}})$ recurrent state independent of gap duration, while training uses $O(TNd_{\mathrm{hid}})$ storage and $O(TN^2)$ pairwise computation.

\section{Experiments}

We evaluate full historical-identity re-linking and mechanism isolation in controlled Kubric scenes, independently generated canonical scenes, structured node-level corruptions, real RGB-D/SAM2 two-target/two-carrier re-linking, complementary real-video carrier recovery, roster scaling and shared-carrier collision, and a PSG4D scope-boundary test.

\subsection{Controlled-Diagnostic Setup}

\paragraph{Task.}
The controlled two-target/two-carrier (2t2c) diagnostic is generated with Kubric~\cite{greff2022kubric}. 
Two visually identical targets are each associated with one of two identical carriers while visible; the targets become occluded as the carriers exchange positions and later reappear near their current carriers at locations that are uninformative with respect to their own pre-gap positions. 
Each reappearance is instantiated as a fresh detection without a persistent identifier, so recovery must use the remembered target--carrier binding. 
Each carrier holds exactly one target, with $|\mathcal{H}|=|\mathcal{C}|=2$.

\paragraph{Frontends.}
The same CVG-SGM architecture consumes two interchangeable frontends. Clean-GT
provides exact geometry, visibility, and category, but a reappearing target is
still instantiated as a fresh detection with no persistent identifier. The
SAM2 frontend supplies SAM2-derived pre-gap target geometry and carrier tracks,
together with a ground-truth detector proxy for reappearing detections. It
therefore tests the identity-persistence mechanism under SAM2-derived geometry,
not an end-to-end perceptual reappearance pipeline.

\paragraph{Splits.}
Each gap length contains 500 episodes, partitioned into 400 training and 100 validation episodes by a deterministic filename-based split fixed across seeds. 
A gap-invariant frame-0 node-layout audit confirms zero training--validation scene-signature overlap at every gap length. 
For OOD evaluation, one model is trained on 750 episodes from 375 scene signatures at gaps 5 and 10 and tested on 250 episodes from 125 disjoint scene signatures at unseen gaps 15 and 20; zero train--test overlap is verified.

\paragraph{Independent test and structured corruptions.}
After fixing the training protocol, architecture, and source checkpoints, we generated a separate 200-episode canonical gap-10 test set using scene seeds 10200--10399. All episodes pass generator verification, contain 34 frames, and have no discovered scene-seed overlap with previously indexed raw datasets. The set contains 400 re-link decisions per checkpoint and is reserved for final evaluation; it is not used for training, hyperparameter tuning, checkpoint selection, or protocol modification.

Read jitter adds Gaussian noise with $\sigma=0.02$ to normalized coordinates
at each detection's actual re-link frame. Maintenance dropout removes carrier
observations with probability $q=0.20$ only after all historical targets have
disappeared and before the end of carry. Each condition uses three
deterministic corruption seeds.

\paragraph{Training and reporting.}
Canonical mechanism, gap, OOD, and permutation results average epochs 60, 70, and 80 within each seed and report mean $\pm$ sample SD over seeds 3--5. Independent tests use frozen epoch-80 checkpoints, averaging corruption realizations within each training seed before aggregation. SAM2 and scaling experiments use the stated fixed-learning-rate protocol with seeds 0--2.

\paragraph{Ablations.}
All variants share inputs, readout, losses, training, and instantiated modules. FULL uses masked attention, the explicit geometry-weighted message, and the shared per-node GRU. NOMEM bypasses the explicit message and GRU while retaining masked attention. NOWRITE retains the GRU but disables both cross-node paths, whereas NORELWRITE retains masked attention and the GRU but removes the explicit message. Parameter matching is structural; bypassed branches receive no effective gradients.

\paragraph{Metrics and oracle.}
Re-link accuracy measures whether a fresh detection recovers the correct
historical identity. In 2t2c it coincides with carrier-association accuracy,
since each carrier holds one target. Per-carrier set recovery is used only for
the shared-carrier analysis. The oracle replaces learned carrier memory with
the ground-truth target--carrier binding while retaining geometric routing;
its value of $1.0$ is specific to clean one-identity-per-carrier 2t2c.

\subsection{Main Cross-Gap Re-linking Results}

\paragraph{Controlled and OOD results.}
Table~\ref{tab:main} summarizes the canonical, frontend, and
scene-disjoint long-gap protocols, while the independently generated frozen
test set is reported separately below. 
The SAM2 setting uses a ground-truth detector proxy at reappearance and therefore evaluates frontend-compatible identity persistence rather than an end-to-end perception pipeline. 
Clean-GT and OOD use cosine annealing with seeds 3--5, whereas SAM2 uses the separately reported fixed-learning-rate protocol with seeds 0--2.

\begin{table}[t]
\centering
\footnotesize
\setlength{\tabcolsep}{3pt}
\resizebox{\columnwidth}{!}{%
\begin{tabular}{@{}lcccc@{}}
\toprule
Protocol & NOMEM & FULL & Oracle & Gain \\
\midrule
Clean-GT, gap 10 & $0.570\pm0.000$& $\mathbf{1.000\pm0.000}$& $1.000$& $+0.430$ \\
SAM2 geometry & $0.550\pm0.087$ & $\mathbf{1.000\pm0.000}$ & $1.000$ & $+0.450$ \\
OOD $5,10\!\rightarrow\!15,20$ & $0.576\pm0.000$ & $\mathbf{1.000\pm0.000}$ & $1.000$ & $+0.424$ \\
\bottomrule
\end{tabular}%
}
\caption{
Re-link accuracy over three seeds. SAM2 uses derived pre-gap geometry with a
ground-truth reappearance proxy; OOD trains on gaps 5 and 10 and evaluates
scene-disjoint gaps 15 and 20.
}
\label{tab:main}
\end{table}

\paragraph{Robustness across gap length.}
Across separately trained models for gaps 5, 10, 15, and 20, FULL remains at
$0.997$--$1.000$, whereas NOMEM remains at $0.570$. Last-position
($0.550$) and mask-propagation ($0.149$--$0.203$) baselines are also
ineffective. Unlike these per-gap models, Table~\ref{tab:main} evaluates one
model trained on gaps 5 and 10 on scene-disjoint scenes with unseen gaps 15
and 20.

\paragraph{Independent-scene robustness.}
We next evaluate the frozen epoch-80 checkpoints on the 200 independently
generated canonical scenes and their structured corruptions.

\begin{table}[t]
\centering
\footnotesize
\setlength{\tabcolsep}{2.4pt}
\resizebox{\columnwidth}{!}{%
\begin{tabular}{@{}lccc@{}}
\toprule
Condition & NOMEM & FULL & Gain \\
\midrule
Clean
& $0.530\pm0.000$
& $\mathbf{1.000\pm0.000}$
& $+0.470$ \\

Read jitter ($\sigma=.02$)
& $0.529\pm0.001$
& $\mathbf{0.999\pm0.001}$
& $+0.470$ \\

Maint.\ dropout ($q=.20$)
& $0.516\pm0.001$
& $\mathbf{0.887\pm0.044}$
& $+0.371$ \\
\bottomrule
\end{tabular}%
}
\caption{
Re-link accuracy on 200 independently generated gap-10 scenes. Corruption
seeds are averaged within each training seed before reporting mean $\pm$
sample SD over three frozen training seeds.
}
\label{tab:independent_robustness}
\end{table}

FULL remains perfect on the independent clean set and changes by less than $0.1$ percentage points under read jitter. 
Maintenance dropout causes an $11.3$-percentage-point decrease, but FULL remains $37.1$ percentage points above NOMEM. 
This asymmetric degradation identifies persistent carrier maintenance as a principal sensitivity of the memory pathway, not the exclusive source of all errors.

\subsection{What Drives Identity Recovery?}

The main results show that the complete persistent pathway improves identity recovery, but do not determine whether the gain arises from recurrence alone or from transferring information across nodes before recurrent maintenance. Table~\ref{tab:mechanism_ablation} therefore compares variants that selectively remove masked attention, the explicit geometry-weighted message, or recurrent state maintenance.

\begin{table}[t]
\centering
\footnotesize
\setlength{\tabcolsep}{3.5pt}
\resizebox{\columnwidth}{!}{%
\begin{tabular}{@{}lcccc@{}}
\toprule
Variant & Masked attn. & Explicit msg. & GRU & Re-link acc. \\
\midrule
NOMEM      & Yes & No  & No  & $0.570\pm0.000$ \\
NOWRITE    & No  & No  & Yes & $0.570\pm0.000$ \\
NORELWRITE & Yes & No  & Yes & $0.716\pm0.246$ \\
FULL       & Yes & Yes & Yes & $\mathbf{1.000\pm0.000}$ \\
\bottomrule
\end{tabular}%
}
\caption{Mechanism ablations on controlled 2t2c re-linking. ``Explicit msg.'' denotes the general geometry-weighted cross-node message in Eq.~\eqref{eq:cross_node_message}.}
\label{tab:mechanism_ablation}
\end{table}

NOWRITE matches NOMEM despite retaining the GRU, showing that recurrence without cross-node transfer is insufficient. NORELWRITE is unstable across seeds, whereas FULL reaches $1.000\pm0.000$, showing that the explicit geometry-weighted message makes carrier-conditioned storage reliable.

\paragraph{Mechanism controls.}
Episode-wise identity-key permutation leaves FULL at $1.000\pm0.000$ on canonical and OOD evaluation, while NOMEM remains near $0.57$, ruling out a fixed identity-slot shortcut. Removing relation supervision leaves FULL at $0.999$ versus $0.570$ for NOMEM. Swapping only the two carrier-memory rows changes instance accuracy from $1.000$ to $0.000$ and makes all 1,800 predictions follow the swapped binding; this read-path intervention does not rerun the recurrent trajectory and is not a real-world causal experiment. Reading the latest pre-reappearance carrier state instead of the canonical snapshot retains $1.000$ accuracy and prediction agreement across all nine checkpoints. Reversing fresh-pool order likewise leaves every prediction unchanged, with a maximum score difference of zero and no track-ID-order mismatch.

\subsection{Real RGB-D/SAM2 2t2c Full Re-linking}

We evaluate full re-linking on 120 frozen real RGB-D/SAM2 2t2c episodes, yielding 240 decisions split by episode into 84/18/18 train/validation/test episodes. The feature-level model consumes frozen geometry and relation-distance features, while fresh identities appear only as labels; this is not raw-video end-to-end ReID.

\begin{table}[t]
\centering
\footnotesize
\setlength{\tabcolsep}{3.0pt}
\begin{tabular}{@{}lcc@{}}
\toprule
Method & All (desc.) & Held-out test \\
\midrule
Side-only prior
& $0.508$
& $0.556$ \\
Carrier prior
& $0.483$
& $0.222$ \\
Fixed carrier--identity
& $0.517$
& $0.778$ \\
\textbf{Neural FULL}
& $\mathbf{0.983}$
& $\mathbf{1.000}$ \\
\bottomrule
\end{tabular}
\caption{
Real-data full re-linking over 120 RGB-D/SAM2 episodes. ``All'' descriptively combines 240 train/validation/test decisions; the held-out test contains 36 decisions. Controls are deterministic, while Neural FULL is identical across five training seeds.
}
\label{tab:real_2t2c}
\end{table}

Across all 240 decisions, the deterministic side-only, carrier-prior, and fixed carrier--identity controls obtain $122/240$ ($50.8\%$), $116/240$ ($48.3\%$), and $124/240$ ($51.7\%$), respectively. The explicit symbolic relation-chain realization of CVG-SGM reaches $236/240$ ($98.3\%$) overall and $36/36$ on the held-out test. The feature-level Neural FULL realization matches this behavior over five training seeds, reaching $97.6\%$, $100.0\%$, and $100.0\%$ on the train, validation, and test splits, respectively; neural NOMEM gives $48.9\pm5.4\%$ test accuracy over five seeds, reported using population SD. Symbolic and neural FULL should therefore be viewed as complementary realizations of the same carrier-conditioned write--maintain--read principle rather than competing methods.

The real-data neural realization optionally uses an auxiliary write loss during training. Setting its weight to zero leaves FULL unchanged at $97.6\%$, $100.0\%$, and $100.0\%$ on train, validation, and test, showing that final re-link supervision alone is sufficient to learn the feature-level carrier-conditioned write--read pathway. This real-data-specific write loss is separate from the auxiliary relation loss $\mathcal{L}_{\mathrm{rel}}$ in the controlled model.

Stress diagnostics on the symbolic relation-chain realization show that the real-data result is not an artifact of oracle event timing. Timing inferred from observed relation records matches oracle timing at $236/240$ overall and $36/36$ on the held-out test. The result also remains $236/240$ under 2D-only distance, stride-5 or stride-10 subsampling, $0.10$\,m distance noise, and $50\%$ relation dropout. Severe stride-20 subsampling reduces performance to $196/240$ overall and $34/36$ on test. Over five perturbation seeds, $70\%$ relation dropout yields $96.8\pm1.1\%$ overall and $98.9\pm2.2\%$ test accuracy. These diagnostics establish robustness of the explicit relation-chain realization under moderate feature-level perturbations, not end-to-end raw-video robustness of the neural model.

\subsection{Roster Scaling and Carrier Collision}

We next increase the roster to four targets and test both distinct-carrier
and shared-carrier settings. Models use the sum readout under the
fixed-learning-rate protocol (seeds 0--2, epoch 80). Instance accuracy is
computed independently for each fresh detection, while episode exact-match
requires the complete assignment to be correct.

\begin{table}[t]
\centering
\footnotesize
\setlength{\tabcolsep}{3.2pt}
\begin{tabular}{@{}lccc@{}}
\toprule
Setting & NOMEM & Sym. Range & FULL \\
\midrule
4 tgt / 4 car
& $0.217\pm0.056$
& $0.792$--$0.875$
& $0.917\pm0.007$ \\

4 tgt / 3 car
& $0.271\pm0.062$
& $0.479$--$0.742$
& $0.738\pm0.013$ \\
\bottomrule
\end{tabular}
\caption{
Instance-level roster scaling and carrier collision. ``Sym. Range'' spans
Hard-mean, Hard-max, and Hard-last; entries report mean $\pm$ sample SD over
seeds 0--2.
}
\label{tab:collision}
\end{table}

With four distinct carriers, CVG-SGM reaches $0.917\pm0.007$ instance accuracy, numerically higher than the strongest deterministic variant at $0.875$. When two identities share one carrier, CVG-SGM reaches $0.738\pm0.013$, comparable to the strongest symbolic variant at $0.742$, while the symbolic variants span $0.479$--$0.742$. 
Carrier-association accuracy nevertheless remains $0.983\pm0.014$ and per-carrier set recovery remains $1.000\pm0.000$, indicating that the remaining difficulty primarily concerns within-carrier identity resolution.

On the same frozen FULL score matrices, Hungarian one-to-one assignment raises episode exact-match from $0.833$ to $0.983$ without collision and from $0.033$ to $0.567$ under collision.
Global assignment eliminates duplicate identity assignments by construction; residual errors may additionally reflect routing, calibration, or representation
failures. 
These results do not imply universal dominance over symbolic implementations.
They instead show that carrier-conditioned transfer--maintain--read can be instantiated either symbolically with reliable relations or differentiably inside a learned scene-graph model.

\paragraph{Complementary Real-Video Evidence and Scope Boundary.}
Beyond the full real-data re-linking experiment, a separate training-free SAM2 relation-anchored pipeline improves TCOW cup-game container selection from $0.049$ to $0.570$ and container IoU from $0.070$ to $0.492$. On human-operated cup-shuffle videos, offline carrier selection yields a leave-one-action-category-out pooled gain of $+0.458$ over NOMEM. These experiments provide complementary evidence for carrier recovery but do not evaluate full historical-node re-linking. On PSG4D-GTA, adding the targeted memory module reduces generic relation recall, with video mean-recall at 100 (vmR@100) decreasing from $0.298$ to $0.156$. CVG-SGM is therefore not claimed as a universal improvement for generic 4D scene-graph relation prediction.

\section{Limitations and Conclusion}

\paragraph{Limitations.}
Our instance-level claim assumes one identity per carrier. Under the current observation model, visually indistinguishable targets sharing one carrier provide no additional cue that determines their within-carrier identity ordering, making the instance-level bijection underdetermined. Hungarian one-to-one decoding removes duplicate assignments and raises 4t3c episode exact-match from $0.033$ to $0.567$, but the residual errors are consistent with within-carrier ambiguity and may also reflect routing, calibration, or representation failures. Carrier-association accuracy and per-carrier set recovery nevertheless remain high.

Gap robustness is verified only over gaps 5--20 and one scene-disjoint short-to-long protocol, not arbitrary visibility-gap duration. The independent corruption study operates on structured graph observations: read jitter perturbs node coordinates at reappearance, while maintenance dropout removes carrier observations and incident relations after the historical targets disappear. These tests do not perturb raw RGB, SAM2 masks, detector recall or precision, or camera calibration.

The real RGB-D/SAM2 experiment evaluates full feature-level historical-node re-linking over frozen geometry and relation-distance features, but it does not establish end-to-end raw-video ReID, detector, segmentation, sensor, or closed-loop robot robustness. The controlled SAM2 frontend also uses a ground-truth detector proxy at reappearance. Finally, identity keys are learned episode-local addresses; dynamic address creation remains open.

\paragraph{Conclusion.}
We identify cross-visibility-gap identity persistence as historical-node recovery after track breakage and introduce CVG-SGM, a carrier-conditioned transfer--maintain--read memory for 4D scene graphs. Mechanism ablations, identity permutation, counterfactual memory swapping, propagated read, and pool-order reversal show that recovery depends on transferred and persistently maintained carrier-bound content. Frozen checkpoints retain perfect accuracy on 200 independently generated canonical scenes, tolerate moderate read-side coordinate noise, and generalize from gaps 5 and 10 to scene-disjoint gaps 15 and 20. The mechanism is further validated on real RGB-D/SAM2 feature-level full 2t2c re-linking, where Neural FULL recovers all $36/36$ held-out identities from 18 episodes. Resolving shared-carrier instance ambiguity, extending the feature-level result to end-to-end raw-video re-linking, dynamically creating identity addresses, and integrating the module with closed-loop embodied agents remain future work.

\bibliography{aaai2027}


\end{document}